\documentclass{article}
\usepackage{spconf,amsmath,graphicx,hyperref}

\usepackage{algorithm}
\usepackage{algorithmic}
\usepackage{bbding} 
\usepackage[normalem]{ulem}
\useunder{\uline}{\ul}{}
\usepackage{amsfonts,amssymb,bm}
\usepackage{balance}
\usepackage{array}
\usepackage[caption=false,font=normalsize,labelfont=sf,textfont=sf]{subfig}
\usepackage{textcomp}
\usepackage{stfloats}
\usepackage{url}
\usepackage{verbatim}
\usepackage{xcolor}
\usepackage{booktabs}
\usepackage{multirow}
\usepackage{enumitem}
\usepackage{makecell}

\usepackage{color}
\usepackage{colortbl}
\usepackage{pifont}
\definecolor{lightblue}{RGB}{221,235,247}
\newcommand{\brow}[1]{\rowcolor{lightblue}}

\definecolor{softred}{rgb}{0.9, 0.4, 0.4}
\definecolor{softgreen}{rgb}{0.4, 0.7, 0.4}
\newcommand{\rtext}[1]{\textbf{\textcolor{softred}{#1}}}
\newcommand{\gtext}[1]{\textbf{\textcolor{softgreen}{#1}}}

\usepackage{xspace}
\newcommand{\methodname}[0]{\textsc{CTcld}\xspace}

\title{Cross-modal Translation via Conditional Latent Denoising for Video Deepfake Detection}
\name{Xinzhe Li, Youzhi Tu, Kong Aik Lee$^{\ast}$ \thanks{$^{\ast}$Corresponding author.}}
\address{Department of Electrical and Electronic Engineering,\\ The Hong Kong Polytechnic University, Hong Kong SAR\\}
\begin{document}
%
\maketitle
\begin{abstract}
The growing threat of video deepfakes necessitates multimodal detection. Beyond serving as independent indicators of authenticity, audio and visual signals have intrinsic dependencies that also provide an essential criterion for detection. Previous methods often overlook the cross-modal correspondences, hindering information transfer between domains and leaving crucial detection cues unexplored. To address this challenge, we propose a framework called Cross-modal Translation via Conditional Latent Denoising (\methodname) for video deepfake detection. It connects the distinct distributions of heterogeneous modalities in latent spaces, enabling smooth cross-domain information transfer to improve detection performance. We first establish a Bayesian foundation by decomposing the audio-visual joint distribution. Subsequently, \methodname translates both modalities via bidirectional latent denoising conditioned on each other, effectively capturing subtle inconsistencies in the manipulated signals. Experimental results demonstrate that the proposed \methodname enables comprehensive domain alignment, resulting in a robust video deepfake detection approach with competitive performance.
\end{abstract}
\begin{keywords}
Multimodal learning, video deepfake detection, audio-visual fusion, latent diffusion models
\end{keywords}
\section{Introduction}
Deepfakes pose profound societal and economic challenges \cite{lisena2026future}. Recently, the rapid development of multimodal deepfakes renders traditional unimodal detection methods ineffective. The strong intrinsic dependencies between audio and visual signals inevitably introduce complex dynamics. Thus, video deepfake detection has emerged as a vital research focus. For instance, \cite{zou2024cross} proposes cross-modality and within-modality regularization to preserve modality-specific representations during audio-visual fusion. \cite{anshul2025next} detects video deepfakes by comparing predicted next-frame features with actual ones through local window attention. However, current approaches struggle to fully exploit crucial detection cues from the cross-modal correspondences. On the one hand, some methods \cite{zhou2021joint,ilyas2023avfakenet} model audio and visual streams independently. On the other hand, some approaches \cite{zhang2024joint,wang2024avt} rely on simple mapping to realize transitions between both domains, which inevitably leads to information degradation.

Since the cross-modal correspondences provide an essential criterion for detection, bridging the gap between heterogeneous modalities is important. Audio and visual signals are distinct in form, but both are manifestations of the same video, thus making the joint modeling in their respective feature spaces challenging. To address this, we propose a multimodal deepfake detection framework called Cross-modal Translation via Conditional Latent Denoising (\methodname), enabling smooth cross-domain information transfer to improve video deepfake detection. We first understand the intrinsic dependencies between audio and visual modalities through the perspective of Bayesian factorization. Thus, the complex audio-visual joint distribution can be mathematically decomposed into a marginal distribution of one modality and a conditional distribution of the other one.

Due to limited information interactions between the two modalities caused by the gap between their original feature spaces, we train audio and visual encoders on authentic videos to obtain modality-specific marginal distributions in the latent space. Furthermore, \methodname models the conditional distributions based on the extracted priors via a bidirectional latent denoising process. Finally, \methodname integrates and leverages them as the source of retrieving detection criteria to isolate the subtle manipulation artifacts characteristic of multimodal deepfakes. Our contributions are summarized as follows:

\begin{itemize}
\item We underscore the importance of the cross-modal correspondences and explore them to provide the essential video deepfake detection criteria. 
\item We propose a framework called \methodname, enabling smooth cross-domain information transfer to improve video deepfake detection based on the decomposition of complex audio-visual joint representation. 
\item Experimental results demonstrate that \methodname better bridges heterogeneous modalities and achieves competitive detection performance across datasets.
\end{itemize}

\section{Methodology}
\subsection{Formulation of Cross-Modal Correspondences}

\begin{figure*}[t]
  \centering
  \includegraphics[width=.99\linewidth]{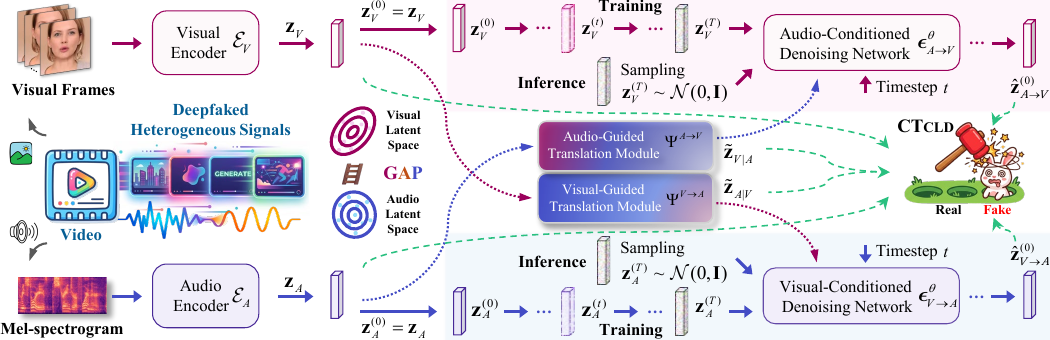}
  \caption{The overall architecture of the Cross-modal Translation via Conditional Latent Denoising (\methodname).}
  \label{fig1}
\end{figure*}
The correspondences between the audio and visual modalities in videos are governed by a complex joint probability distribution $p(\mathbf{z}_A, \mathbf{z}_V)$, where $\mathbf{z}_A$ and $\mathbf{z}_V$ denote the latent audio and visual embeddings, respectively. However, directly estimating $p(\mathbf{z}_A, \mathbf{z}_V)$ is computationally cost-intensive due to the gap between the heterogeneous modalities. Considering the bidirectional nature of audio-visual correspondences, we decompose the joint probability distribution via a symmetrized Bayesian factorization:
\begin{equation}
    \begin{aligned}
    p(\mathbf{z}_A, \mathbf{z}_V) = \left[ \, p(\mathbf{z}_V | \mathbf{z}_A) \, p(\mathbf{z}_A | \mathbf{z}_V) \, p(\mathbf{z}_V) \, p(\mathbf{z}_A) \, \right]^{\frac{1}{2}},
    \end{aligned}
\end{equation}
where $p(\mathbf{z}_A)$ and $p(\mathbf{z}_V)$ are the modality-specific marginal distributions which are extracted by audio and visual encoders $\mathcal{E}_A$ and $\mathcal{E}_V$ in the latent space. $p(\mathbf{z}_V | \mathbf{z}_A)$ and $p(\mathbf{z}_A | \mathbf{z}_V)$ are the cross-modal conditional distributions. Building upon this formulation, we perform translations between domains based on the underlying mechanisms of cross-modal correspondences.

\subsection{Cross-Modal Translation}
As illustrated in Fig.~\ref{fig1}, visual frames and a Mel-spectrogram are first fed into their respective encoders $\mathcal{E}_V$ and $\mathcal{E}_A$ to extract the modality-specific latent embeddings $\mathbf{z}_V$ and $\mathbf{z}_A$. Within the visual latent space $\mathcal{M}_V$, an audio-guided translation module $\Psi^{A \to V}$ is employed to project the audio representation $\mathbf{z}_A$ into an audio-conditioned visual prior $\tilde{\mathbf{z}}_{V|A}$. This prior subsequently serves as the conditioning signal during a latent denoising process to guide the generative reconstruction of the visual embeddings $\mathbf{z}_V$. Similarly, a symmetrical translation is simultaneously performed for the audio modality, guided by the visual-conditioned audio prior $\tilde{\mathbf{z}}_{A|V}$. Finally, \methodname integrates these enriched representations to systematically expose forgery artifacts, serving as the source for the video deepfake detection. Specifically, we denote $\mathbf{z}_V^{(0)} := \mathbf{z}_V \in \mathcal{M}_V$  and $\mathbf{z}_A^{(0)} := \mathbf{z}_A \in \mathcal{M}_A$ as the clean visual and audio latent representations. A Markovian corruption chain over $T$ timesteps progressively injects Gaussian noise into $\mathbf{z}_V^{(0)}$ according to a linear variance schedule $\beta_1, \dots, \beta_T$:
\begin{equation}
    \resizebox{.91\columnwidth}{!}{ $
    q\left(\mathbf{z}_V^{(1:T)} \;\middle|\; \mathbf{z}_V^{(0)}\right) = \prod_{t=1}^T \mathcal{N}\left(\mathbf{z}_V^{(t)}; \; \sqrt{1 - \beta_t}\,\mathbf{z}_V^{(t-1)}, \; \beta_t \mathbf{I}\right).
    $ }
\end{equation}
Defining $\alpha_t = 1 - \beta_t$ and $\bar{\alpha}_t = \prod_{s=1}^t \alpha_s$, the marginal distribution at any arbitrary diffusion timestep $t \in \{1, \dots, T\}$ admits a closed-form Gaussian formulation:
\begin{equation}
    q\left(\mathbf{z}_V^{(t)} \;\middle|\; \mathbf{z}_V^{(0)}\right) = \mathcal{N}\left(\mathbf{z}_V^{(t)}; \; \sqrt{\bar{\alpha}_t}\,\mathbf{z}_V^{(0)}, \; (1 - \bar{\alpha}_t)\mathbf{I}\right).
\end{equation}
To provide domain-aligned conditioning, we formulate the audio-conditioned visual prior as $\tilde{\mathbf{z}}_{V|A}:= \Psi^{A \to V}(\mathbf{z}_A)$. The reverse process reconstructs the visual latent representation from Gaussian noise $\mathbf{z}_V^{(T)} \sim \mathcal{N}(\mathbf{0}, \mathbf{I})$, conditioned explicitly on $\tilde{\mathbf{z}}_{V|A}$. We parameterize the reverse Markov chain as:
\begin{equation}
    \resizebox{.91\columnwidth}{!}{ $
    p_\theta\left(\mathbf{z}_V^{(0:T)} \;\middle|\; \tilde{\mathbf{z}}_{V|A}\right) = p\left(\mathbf{z}_V^{(T)}\right) \prod_{t=1}^T p_\theta\left(\mathbf{z}_V^{(t-1)} \;\middle|\; \mathbf{z}_V^{(t)}, \tilde{\mathbf{z}}_{V|A}\right), 
    $ }
\end{equation}
\begin{equation}
    \resizebox{.9\columnwidth}{!}{ $
    \begin{aligned}
        &p_\theta\left(\mathbf{z}_V^{(t-1)} \;\middle|\; \mathbf{z}_V^{(t)}, \tilde{\mathbf{z}}_{V|A}\right) \\
        &= \mathcal{N}\left(\mathbf{z}_V^{(t-1)}; \; \boldsymbol{\mu}_\theta\left(\mathbf{z}_V^{(t)}, t, \tilde{\mathbf{z}}_{V|A} \right), \; \mathbf{\Sigma}_\theta\left(\mathbf{z}_V^{(t)}, t, \tilde{\mathbf{z}}_{V|A}\right)\right). 
    \end{aligned}
    $ }
\end{equation}

Since directly maximizing the $\log p_\theta(\mathbf{z}_V^{(0)} | \tilde{\mathbf{z}}_{V|A})$ is intractable, we instead optimize its variational evidence lower bound. By introducing the forward diffusion trajectory $q(\mathbf{z}_V^{(1:T)} | \mathbf{z}_V^{(0)})$ and applying Jensen's inequality, the log-likelihood is bounded as follows:
\begin{equation}
    \resizebox{0.91\columnwidth}{!}{$
    \begin{aligned}
        \log p_\theta\left(\mathbf{z}_V^{(0)} \;\middle|\; \tilde{\mathbf{z}}_{V|A}\right) &= \log \mathbb{E}_{q\left(\mathbf{z}_V^{(1:T)} \mid \mathbf{z}_V^{(0)}\right)} \left[ \frac{p_\theta\left(\mathbf{z}_V^{(0:T)} \;\middle|\; \tilde{\mathbf{z}}_{V|A}\right)}{q\left(\mathbf{z}_V^{(1:T)} \;\middle|\; \mathbf{z}_V^{(0)}\right)} \right] \\
        &\ge \mathbb{E}_{q} \left[ \log \frac{p_\theta\left(\mathbf{z}_V^{(0:T)} \;\middle|\; \tilde{\mathbf{z}}_{V|A} \right)}{q\left(\mathbf{z}_V^{(1:T)} \;\middle|\; \mathbf{z}_V^{(0)}\right)} \right] := -\mathcal{L}_{\text{VLB}}.
    \end{aligned}
    $}
\end{equation}
By factorizing the joint distributions via the Markov property, $\mathcal{L}_{\text{VLB}}$ can be decomposed into a sum of Kullback-Leibler (KL) divergences and a reconstruction term:
\begin{equation}
    \resizebox{0.91\columnwidth}{!}{$
    \begin{aligned}
        \mathcal{L}_{\text{VLB}} &= \mathbb{E}_{q} \Bigg[ D_{\text{KL}}\left( q\left(\mathbf{z}_V^{(T)} \;\middle|\; \mathbf{z}_V^{(0)}\right) \;\big\|\; p\left(\mathbf{z}_V^{(T)}\right) \right) \\
        &\quad + \sum_{t=2}^T D_{\text{KL}}\left( q\left(\mathbf{z}_V^{(t-1)} \;\middle|\; \mathbf{z}_V^{(t)}, \mathbf{z}_V^{(0)}\right) \;\big\|\; p_\theta\left(\mathbf{z}_V^{(t-1)} \;\middle|\; \mathbf{z}_V^{(t)}, \tilde{\mathbf{z}}_{V|A} \right) \right) \\
        &\quad - \log p_\theta\left(\mathbf{z}_V^{(0)} \;\middle|\; \mathbf{z}_V^{(1)}, \tilde{\mathbf{z}}_{V|A}\right) \Bigg].
    \end{aligned}
        $}
\end{equation}
Since the forward schedule $\beta_t$ is fixed and both the forward process posterior and the transitions of the reverse process are Gaussian, minimizing the KL divergence simplifies to matching their mean vectors. By reparameterizing $\mathbf{z}_V^{(0)}$, the mean of the forward process posterior $\tilde{\boldsymbol{\mu}}_t$ can be expressed as an affine combination of the current noisy state $\mathbf{z}_V^{(t)}$ and the injected noise $\boldsymbol{\epsilon}$:
\begin{equation}
    \resizebox{.7\columnwidth}{!}{ $
    \tilde{\boldsymbol{\mu}}_t\left(\mathbf{z}_V^{(t)}, \mathbf{z}_V^{(0)}\right) = \frac{1}{\sqrt{\alpha_t}} \left( \mathbf{z}_V^{(t)} - \frac{\beta_t}{\sqrt{1 - \bar{\alpha}_t}} \boldsymbol{\epsilon} \right).
    $ }
\end{equation}
During the reverse process, since $\mathbf{z}_V^{(t)}$ is explicitly available, estimating $\tilde{\boldsymbol{\mu}}_t$ mathematically reduces to predicting the injected noise $\boldsymbol{\epsilon}$. Thus, we parameterize an audio-conditioned denoising network $\boldsymbol{\epsilon}_{A \to V}^\theta$. The resulting audio-to-visual denoising objective $\mathcal{L}_{A \to V}^{\text{diff}}$ is formulated as:
\begin{equation}
\mathcal{L}_{A \to V}^{\text{diff}} = \mathbb{E}_{\mathbf{z}_V^{(0)}, \tilde{\mathbf{z}}_{V|A} , \boldsymbol{\epsilon}_V \sim \mathcal{N}(\mathbf{0}, \mathbf{I}), t} \, \left[ \ell_{A \to V}^t \right],
\end{equation}
\begin{equation}
    \resizebox{.89\columnwidth}{!}{ $
    \ell_{A \to V}^t = \left\| \boldsymbol{\epsilon}_V - \boldsymbol{\epsilon}_{A \to V}^\theta\left( \sqrt{\bar{\alpha}_t}\mathbf{z}_V^{(0)} + \sqrt{1-\bar{\alpha}_t}\,\boldsymbol{\epsilon}_V, \; t, \; \tilde{\mathbf{z}}_{V|A}  \right) \right\|_2^2.
    $ }
\end{equation}
Symmetrically, the visual-conditioned denoising network $\boldsymbol{\epsilon}_{V \to A}^\theta$ is optimized via the corresponding visual-to-audio denoising objective $\mathcal{L}_{V \to A}^{\text{diff}}$, which is defined as follows:
\begin{equation}
\mathcal{L}_{V \to A}^{\text{diff}} = \mathbb{E}_{\mathbf{z}_A^{(0)}, \tilde{\mathbf{z}}_{A|V} , \boldsymbol{\epsilon}_A \sim \mathcal{N}(\mathbf{0}, \mathbf{I}), t} \, \left[ \ell_{V \to A}^t \right],
\end{equation}
\begin{equation}
    \resizebox{.89\columnwidth}{!}{ $
    \ell_{V \to A}^t = \left\| \boldsymbol{\epsilon}_A - \boldsymbol{\epsilon}_{V \to A}^\theta\left( \sqrt{\bar{\alpha}_t}\mathbf{z}_A^{(0)} + \sqrt{1-\bar{\alpha}_t}\,\boldsymbol{\epsilon}_A, \; t, \; \tilde{\mathbf{z}}_{A|V}  \right) \right\|_2^2.
    $ }
\end{equation}

Specifically, we first fuse the audio-conditioned visual prior $\tilde{\mathbf{z}}_{V|A}$ with the noisy visual state $\mathbf{z}_V^{(t)}$ via a token-wise additive projection to obtain the initial hidden state $\mathbf{H}^{(0)}_V = \mathbf{W}_v \mathbf{z}_V^{(t)} + \mathbf{W}_{v|a} \tilde{\mathbf{z}}_{V|A}$, where $\mathbf{W}_v$ and $\mathbf{W}_{v|a}$ are learnable linear projection weights. The fused cross-modal representation is then processed through $K$ cascaded diffusion transformer blocks \cite{peebles2023scalable}. The diffusion timestep $t$ is embedded via a multi-layer perceptron to yield the dimension-wise affine modulation parameters $[\boldsymbol{\gamma}_1^{(k)}, \boldsymbol{\beta}_1^{(k)}, \boldsymbol{\alpha}_1^{(k)}, \boldsymbol{\gamma}_2^{(k)}, \boldsymbol{\beta}_2^{(k)}, \boldsymbol{\alpha}_2^{(k)}]$ for each block. These parameters adaptively modulate the hidden states as follows:
\begin{equation}
\resizebox{.89\columnwidth}{!}{$
\tilde{\mathbf{H}}^{(k)} = \mathbf{H}^{(k-1)} + \boldsymbol{\alpha}_1^{(k)} \odot \text{MHSA}\Big( (\mathbf{1} + \boldsymbol{\gamma}_1^{(k)}) \odot \text{LN}(\mathbf{H}^{(k-1)}) + \boldsymbol{\beta}_1^{(k)} \Big),
$}
\end{equation}
\begin{equation}
\resizebox{.83\columnwidth}{!}{$
\mathbf{H}^{(k)} = \tilde{\mathbf{H}}^{(k)} + \boldsymbol{\alpha}_2^{(k)} \odot \text{FFN}\Big( (\mathbf{1} + \boldsymbol{\gamma}_2^{(k)}) \odot \text{LN}(\tilde{\mathbf{H}}^{(k)}) + \boldsymbol{\beta}_2^{(k)} \Big),
$}
\end{equation}
where $\mathbf{H}^{(k-1)}$, $\tilde{\mathbf{H}}^{(k)}$, and $\mathbf{H}^{(k)}$ denote the input, intermediate, and output hidden states of the $k$-th block, respectively. $\text{LN}\left(\cdot\right)$, $\text{MHSA}\left(\cdot\right)$, and $\text{FFN}\left(\cdot\right)$ represent layer normalization, multi-head self-attention, and the feed-forward network. $\mathbf{1}$ is an all-ones vector, and the operator $\odot$ denotes the element-wise Hadamard product. This adaptive mechanism enables the network to prioritize global cross-modal structural alignment at high noise scales ($t \to T$) while delicately refining localized audio-visual correspondences at low noise scales ($t \to 0$). Finally, the output of the last block is projected back to the input latent dimension to yield the estimated noise:
\begin{equation}
    \hat{\boldsymbol{\epsilon}} = \mathbf{W}_{\text{out}} \, \text{LN}(\mathbf{H}^{(K)}),
\end{equation}
where $\mathbf{W}_{\text{out}}$ is a learnable projection matrix. \methodname leverages feature-level representations by concatenating embeddings along the channel dimension. To discriminate between real and fake videos, a detection head $\mathcal{D}$ is designed to output predicted forgery score $s_{\text{fake}}$:
\begin{equation}
    s_{\text{fake}} = \mathcal{D} \Big( \mathbf{z}_V^{(0)} \oplus \tilde{\mathbf{z}}_{V|A} \oplus \hat{\mathbf{z}}_{A \to V}^{(0)}, \; \mathbf{z}_A^{(0)} \oplus \tilde{\mathbf{z}}_{A|V} \oplus \hat{\mathbf{z}}_{V \to A}^{(0)} \Big),
\end{equation}
where $\oplus$ denotes the channel-wise concatenation operator. $\hat{\mathbf{z}}_{A \to V}^{(0)}$ and $\hat{\mathbf{z}}_{V \to A}^{(0)}$ represent the final reconstructed visual and audio latent representations at timestep $t=0$, conditioned on their corresponding cross-modal priors. Thus, the overall objective function is defined as:
\begin{equation}
    \mathcal{L}_{\text{total}} = \mathcal{L}_{\text{BCE}}(s_{\text{fake}}, y) + \lambda \left( \mathcal{L}_{A \to V}^{\text{diff}} + \mathcal{L}_{V \to A}^{\text{diff}} \right),
\end{equation}
where $\mathcal{L}_{\text{BCE}}\left(\cdot,\cdot\right)$ denotes the binary cross-entropy loss, $y\in \{0, 1\}$ is the ground-truth label, and $\lambda$ is a hyperparameter balancing forgery detection and cross-modal reconstruction to expose audio-visual inconsistencies.

\section{Experiments}
\subsection{Implementation Details}
We uniformly sample 16 visual frames from a clip duration of 3.2s with a resizing operation to map them into a fixed spatial resolution $\mathbb{R}^{224 \times 224}$. Also, we resample the audio to 16 kHz and convert it to Mel-spectrograms of 128 mel-frequency bins with a 20 ms Hann window every 4 ms, where the temporal dimension is 768. Our method is trained on an NVIDIA GeForce RTX 5090 GPU. We first pre-train the encoders $\mathcal{E}_V$ and $\mathcal{E}_A$ on the VoxCeleb2 dataset \cite{chung2018voxceleb2} to obtain $\mathbf{z}_V$ and $\mathbf{z}_A$. We further train our \methodname on FakeAVCeleb \cite{khalid2021fakeavceleb} and DeepFake Detection Challenge (DFDC) \cite{dolhansky2020deepfake} datasets. The Accuracy (Acc) and Area Under Curve (AUC) are utilized as the metrics in evaluations. We employ the denoising diffusion implicit models \cite{SongME21} to accelerate the sampling process. The diffusion loss weight $\lambda$ and the total number of training timesteps and sampling steps are set to $10^{-2}$, 1000, and 5, respectively.

\subsection{Overall Comparison and Ablation Study}
\begin{table}[t]
    \centering
    \caption{Performance comparison results of video deepfake detection across datasets, where \textit{U.} and \textit{M.} denote unimodal and multimodal methods, respectively. \textbf{Bold} and \underline{underlined} numbers are the best and second-best performance.}
    \resizebox{\columnwidth}{!}{
    \begin{tabular}{lc|cccc}
    \toprule
    \multicolumn{2}{c}{\multirow{2}{*}[-0.5ex]{\textbf{Method}}} & \multicolumn{2}{c}{\textbf{FakeAVCeleb}} & \multicolumn{2}{c}{\textbf{DFDC}}
    \\ \cmidrule(lr){3-4} \cmidrule(lr){5-6}
    & & Acc$\uparrow$ (\%) & AUC$\uparrow$ (\%) & Acc$\uparrow$ (\%) & AUC$\uparrow$ (\%)\\
    \midrule
    \multirow{3}{*}{\textit{U.}} & Xception \cite{rossler2019faceforensics++} & 67.9 & 70.5 & 80.5 & 79.3\\
    &  LipForensics \cite{haliassos2021lips} & 80.1 & 82.4 & 71.3 & 73.5\\
    &  RealForensics \cite{haliassos2022leveraging} & 90.1 & 92.3 & 89.6 & 91.5\\
    \midrule
    \multirow{10}{*}{\textit{M.}} & Joint-AVD \cite{zhou2021joint} & 82.5 & 83.3 & 90.2 & 91.9\\
    &  AVFakeNet \cite{ilyas2023avfakenet} & 78.4 & 83.4 & 82.8 & 86.2\\
    &  MCL \cite{liu2023mcl} & 86.0 & 89.3 & \underline{97.9} & 98.3\\
    &  VFD \cite{cheng2024voice} & 81.5 & 86.1 & 81.0 & 85.1\\
    &  PVASS-MDD \cite{yu2024pvass} & 95.7 & 97.3 & 96.3 & \underline{98.9}\\
    &  AVFF \cite{oorloff2024avff} & \underline{98.6} & \underline{99.1} & N/A & N/A\\
    &  AVT$^2$-DWF \cite{wang2024avt} & 87.6 & 88.3 & 88.0 & 89.2\\
    &  AVA-CL \cite{zhang2024joint} & 86.6 & 89.5 & 84.2 & 88.6\\
    &  NFFP \cite{anshul2025next} & 94.8 & 96.3 & N/A & N/A\\
    \cmidrule{2-6}
    \brow & &  \textbf{\methodname (Ours)} & \textbf{99.4} & \textbf{99.5} & \textbf{98.5} & \textbf{99.9}\\
    \bottomrule
    \end{tabular}
    }
    \label{tab:table1}
\end{table}

In Table~\ref{tab:table1}, we compare \methodname with both unimodal and multimodal methods, where \methodname achieves the best performance. On the FakeAVCeleb dataset, \methodname yields 99.4\% in Acc and 99.5\% in AUC, outperforming the second-best method by 0.8\% and 0.4\%, respectively. On the DFDC dataset, \methodname achieves 98.5\% in Acc and 99.9\% in AUC. These results demonstrate that \methodname effectively magnifies cross-modal inconsistencies and exposes subtle audio-visual discrepancies. Furthermore, Table~\ref{tab:table2} validates the necessity of bidirectional cross-modal translation by separately removing the audio- or visual-conditioned latent denoising modules. Specifically, the Acc and AUC results of Models 1 and 2 both drop. The degradation in both unidirectional settings confirms that bidirectional translation captures complementary cross-modal inconsistencies essential for deepfake detection.

\begin{table}[t]
  \centering
  \caption{Results of the ablation study on the FakeAVCeleb.}
  \resizebox{\columnwidth}{!}{
    \begin{tabular}{ccccc}
  \toprule
  \textbf{Model} & \textbf{\makecell{Audio-Conditioned \\ Latent Denoising}} & \textbf{\makecell{Visual-Conditioned \\ Latent Denoising}} & Acc$\uparrow$ (\%) & AUC$\uparrow$ (\%) \\ \midrule
  1 & \gtext{\checkmark} & \rtext{$\times$}  & 97.0 & 92.6 \\
  2 & \rtext{$\times$} & \gtext{\checkmark}  & 98.0 & 97.7 \\
  \textbf{Full} & \gtext{\checkmark} & \gtext{\checkmark} & \textbf{99.4} & \textbf{99.5} \\
  \bottomrule
  \end{tabular}
  }
  \label{tab:table2}
\end{table}

\subsection{Visualization and Sensitivity Analysis}
\begin{figure}[t]
  \centering
  \includegraphics[width=.81\linewidth]{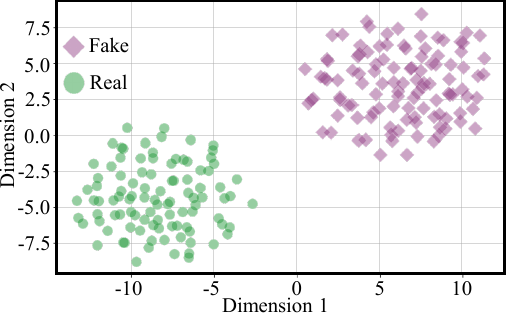}
  \caption{t-SNE visualization of the learned representations used for video deepfake detection.}
  \label{fig2}
\end{figure}

\begin{figure}[t]
  \centering
  \includegraphics[width=.8\linewidth]{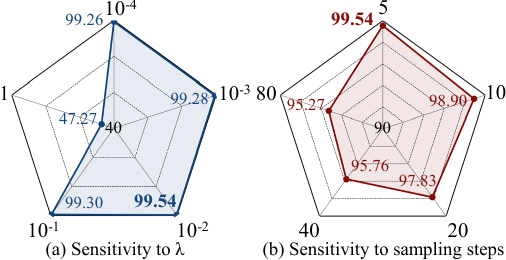}
  \caption{AUC sensitivity to loss weight $\lambda$ and sampling steps.}
  \label{fig3}
\end{figure}

To qualitatively validate the discriminative capability of the learned representations, we visualize the latent feature space after concatenation in \methodname using t-Distributed Stochastic Neighbor Embedding (t-SNE) \cite{van2008visualizing}. Fig.~\ref{fig2} demonstrates that \methodname reveals a clear separation between representations of real and fake videos. Furthermore, we conduct sensitivity analysis on hyperparameters in \methodname. As shown in Fig.~\ref{fig3} (a), we vary the loss weight $\lambda$ across $\{10^{-4}, 10^{-3}, 10^{-2}, 10^{-1}, 1\}$. Competitive results are consistently achieved when setting $\lambda$ within a reasonable range, proving the efficacy of $\mathcal{L}_{A \to V}^{\text{diff}}$ and $\mathcal{L}_{V \to A}^{\text{diff}}$. Also, excessive weighting degrades performance due to the inappropriate shift of the optimization focus. As shown in Fig.~\ref{fig3} (b), we investigate how the denoising trajectory affects the model's sensitivity to subtle manipulation artifacts. We conduct a sensitivity analysis on the sampling steps $\in \{5, 10, 20, 40, 80\}$. Detection performance consistently decreases as the number of sampling steps increases. This trend demonstrates that a denser generative process tends to overly pay attention to manipulation artifacts and reconstruct them, whereas coarse-grained sampling effectively preserves them.


\section{Conclusion}
We have proposed \methodname, a method called Cross-modal Translation via Conditional Latent Denoising, to capture subtle multimodal inconsistencies in video deepfakes. Our framework addresses the challenge of inadequately modeling complex cross-modal correspondences by bridging the gap between heterogeneous modalities in the latent space. Experimental results demonstrate the competitive detection performance of \methodname, achieving Acc and AUC scores of 99.4\% and 99.5\% on the FakeAVCeleb dataset, and 98.5\% and 99.9\% on the DFDC dataset, respectively. Also, we envision this generative translation paradigm inspiring broader applications in other multimodal signal processing.

\bibliographystyle{IEEEbib}
\bibliography{refs}

\end{document}